\documentclass[11pt]{article}
\usepackage{graphicx}
\usepackage{longtable,lscape}
\usepackage{amsthm}
\usepackage{amsfonts}
\usepackage{amsmath}
\usepackage{bbm}
\usepackage{float}
\usepackage{url}

\newcommand{\captionfonts}{\footnotesize}
\makeatletter  
\long\def\@makecaption#1#2{%
  \vskip\abovecaptionskip
  \sbox\@tempboxa{{\captionfonts #1: #2}}%
  \ifdim \wd\@tempboxa >\hsize
    {\captionfonts #1: #2\par}
  \else
    \hbox to\hsize{\hfil\box\@tempboxa\hfil}%
  \fi
  \vskip\belowcaptionskip}
\makeatother 
\begin{document}
\title{Entanglement Zoo II: Examples in Physics and Cognition}
\author{Diederik Aerts and Sandro Sozzo
		\vspace{0.5 cm} \\ 
        \normalsize\itshape
        Center Leo Apostel for Interdisciplinary Studies \\
        \normalsize\itshape
        and, Department of Mathematics, Brussels Free University \\ 
        \normalsize\itshape
         Krijgskundestraat 33, 1160 Brussels, Belgium \\
        \normalsize
        E-Mails: \url{diraerts@vub.ac.be,ssozzo@vub.ac.be}
    		 \\
              }
\date{}
\maketitle
\begin{abstract}
\noindent
We have recently presented a general scheme enabling quantum modeling of different types of situations that violate Bell's inequalities \cite{asIQSA2012}. In this paper, we specify this scheme for a combination of two concepts. We work out a quantum Hilbert space model where `entangled measurements' occur in addition to the expected `entanglement between the component concepts', or `state entanglement'. We extend this result to a macroscopic physical entity, the `connected vessels of water', which maximally violates Bell's inequalities. We enlighten the structural and conceptual analogies between the cognitive and physical situations which are both examples of a nonlocal non-marginal box modeling in our
classification.
\end{abstract}

\section{Introduction\label{intro}}
The presence of entanglement in microscopic quantum particles is typically revealed by a violation of Bell-type inequalities \cite{bell1964,chsh69}. Such a violation also implies that the corresponding coincidence measurements contain correlations that cannot be modeled in a classical Kolmogorovian probability structure \cite{aerts1986,af1982,pitowsky1989}, which led to the widespread belief that such correlations only appear in the micro-world. 
Many years ago, we already showed that Bell's inequalities can be violated by macroscopic physical entities, e.g., two connected vessels of water \cite{aerts1982,aerts1985a,aerts1985b,aerts1991,aertsaertsbroekaertgabora2000}. Little attention was paid to this result at that time, however, because most quantum foundations physicists were convinced that it was impossible to violate Bell's inequalities in situations pertaining to domains different from the micro-world. 
Now that quantum interaction research is flourishing \cite{aertsaerts95,aertsgabora2005a,aertsgabora2005b,bruzaetal2007,bruzaetal2008,aerts2009,bruzaetal2009,pb2009,k2010,songetal2011,bpft2011,bb2012,busemeyeretal2012,ags2012,abgs2012}, it is valuable to reconsider some of these examples, also because we have recently found that it is possible to build Hilbert space models for them, something we did not look into at the time. The fact that it is possible to explicitly construct complex Hilbert space models for these situations became clear to us when we were struggling to quantum-model 
the experimental correlation experiments we performed on a conceptual combination {\it The Animal Acts}.
The Hilbert space modeling of our cognitive correlation data produced a range of new insights. After we had verified that the given concept combination violated Bell's inequalities \cite{ags2012,abgs2012,as2011}, the elaboration of a Hilbert space representation showed the presence of `conceptual entanglement' and proved that this entanglement is only partly due to the component concepts, or `state entanglement', because it is also caused by `entangled measurements' and `entangled dynamical evolutions between measurements' \cite{asIQSA2012}. This discovery of the presence of entanglement on the level of measurements 
and evolutions shed unexpectedly 
new light on traditional in-depth studies of aspects of entanglement, such as the possible violation of the marginal distribution law. 

We have since developed a general quantum modeling scheme for the structural description of the entanglement present in different types of situations violating Bell's inequalities \cite{asQI2013}. In this perspective, situations are possible in which only states are entangled and measurements are products (`customary entanglement'), 
but also situations in which 
entanglement appears on the level of the measurements, in the form of the presence of entangled measurements and the presence of entangled 
evolutions (`nonlocal box situation', `nonlocal non-marginal box situation'). In the present paper, after briefly resuming our empirical results on {\it The Animal Acts} (Sec. \ref{animalactsdescription}), we provide a synthetized 
version of our quantum-theoretic modeling for this conceptual entity (Sec. \ref{animalactsquantum}). We then build a quantum model in complex Hilbert space for the entity `vessels of water' (Secs. \ref{vesselsdescription} and \ref{vessels}). 
This modeling in Hilbert space was not expected when this example was originally conceived, and 
hence constitutes a new result. We also study the conceptual and structural connections between these two situations in the light of our classification in Ref. \cite{asQI2013}. The two cases we consider here are paradigmatic of `nonlocal non-marginal box situations', that is, experimental situations in which (i) joint probabilities do not factorize, (ii) Bell's inequalities are violated, and (iii) the marginal distribution law does not hold. Whenever these conditions are simultaneously satisfied, a form of entanglement appears which is stronger than the `customarily identified quantum entanglement in the states of microscopic entities'. In these cases, it is not possible to work out a quantum-mechanical representation in
a fixed ${\mathbb C}^2\otimes{\mathbb C}^2$ space which satisfies empirical data and where only the initial state is entangled while the measurements are products. It follows that 
entanglement is 
a more complex property than usually thought, a situation we investigate in depth in \cite{asIQSA2012}. 
Shortly, if a single measurement is 
at play, one can distribute the entanglement between state and measurement, but if more measurements are 
considered, the marginal distribution law imposes drastic limits on 
the ways to model the presence of the entanglement \cite{asQI2013}. This is explicitly shown by constructing an alternative ${\mathbb C}^{4}$ modeling for the original vessels of water example (Sec. \ref{alternativevessels}).

Let us remark that we use the naming `entanglement' referring explicitly to the structure within the theory of quantum physics that a modeling of experimental data takes, if (i) these data are represented, following carefully the rules of standard quantum theory, in a complex Hilbert space, and hence states, measurements, and evolutions, are presented respectively by vectors (or density operators), self-adjoint operators, and unitary operators in this Hilbert space; (ii) a situation of coincidence joint measurement on a compound entity is considered, and the subentities are identified following the tensor product rule of `compound entity description in quantum theory' (iii) within this tensor product description of the compound entity entanglement is identified, as `not being product', whether it is for states (non-product vectors), measurements (non-product self-adjoint operators), or evolutions (non-product unitary transformations).

Let us also remark that the research we present in this paper frames within the general emergence of `quantum interaction research'. Indeed, there is increasing evidence 
that quantum structures are systematically present in domains other than the micro-world described by quantum physics. Cognitive science, economics, biology, computer science - they all entail situations that can be modeled more faithfully by
elements of quantum theory than by approaches rooted in classical theories such as classical probability \cite{aertsaerts95,aertsgabora2005a,aertsgabora2005b,bruzaetal2007,bruzaetal2008,aerts2009,bruzaetal2009,pb2009,k2010,songetal2011,bpft2011,bb2012,busemeyeretal2012,ags2012,abgs2012}. Our inspiration to identify quantum structures in domains different from the micro-world originally arose when we were investigating the structures of classical and quantum probability, more specifically, when we were analyzing the question of whether classical probability can reproduce the predictions of quantum theory \cite{aerts1986,aerts1999b}. Understanding the structural difference between classical and quantum probability led us firstly to identify situations in the macroscopic world entailing aspects that are usually attributed only to microscopic quantum entities, such as `contextuality', `emergence', `entanglement, `interference' and `superposition' \cite{aerts1982,aerts1985a,aerts1985b,aerts1991,aertsaertsbroekaertgabora2000}. Later, we extended our search to the realm of human cognition, the structure of human decision processes and the way in which the human mind handles concepts, their dynamics and combinations \cite{aertsaerts95,aertsgabora2005a,aertsgabora2005b,aerts2009,aerts2009b,as2011}. 

\section{The Animal Acts and its quantum representation\label{animalacts}}
We have recently performed a cognitive test on the combination of concepts {\it The Animal Acts} \cite{ags2012,abgs2012,as2011} which violated Bell's inequalities. We have also worked out a quantum representation which fits the collected data and reveals entanglement between the component concepts 
{\it Animal} and {\it Acts}. And, more, it 
shows a `stronger form of entanglement' involving not only entangled states but also entangled measurements
and entangled 
evolutions \cite{asIQSA2012}. In the following, we present these results in the light of the 
classification elaborated in Ref. \cite{asQI2013}.

\subsection{Description of the cognitive test\label{animalactsdescription}}
We consider the sentence {\it The Animal Acts} as a conceptual combination of the concepts {\it Animal} and {\it Acts}. Measurements consists of asking participants in the experiment to answer the question whether a given exemplar `is a good example' of the considered concept or conceptual combination. The measurement $A$, respectively $A'$, considers the exemplars {\it Horse} and {\it Bear}, respectively {\it Tiger} and {\it Cat}, of the concept {\it Animal}, the measurement $B$, respectively $B'$, considers the exemplars {\it Growls} and {\it Whinnies}, respectively {\it Snorts} and {\it Meows}, of the concept {\it Acts}. For the coincidence experiments, for $AB$, participants choose among the four possibilities (1) {\it The Horse Growls}, (2) {\it The Bear Whinnies} -- and if one of these is chosen we put $\lambda_{A_1B_1}=\lambda_{A_2B_2}=+1$ -- and (3) {\it The Horse Whinnies}, (4) {\it The Bear Growls} -- and if one of these is chosen we put $\lambda_{A_1B_2}=\lambda_{A_2B_1}=-1$. For the measurement $AB'$, they choose among (1) {\it The Horse Snorts}, (2) {\it The Bear Meows} -- and in case one of these is chosen we put $\lambda_{A_1B'_1}=\lambda_{A_2B'_2}=+1$ -- and (3) {\it The Horse Meows}, (4) {\it The Bear Snorts} -- and in case one of these is chosen we put $\lambda_{A_1B'_2}=\lambda_{A_2B'_1}=-1$. For the measurement $A'B$, they choose among (1) {\it The Tiger Growls}, (2) {\it The Cat Whinnies} -- and in case one of these is chosen we put $\lambda_{A'_1B_1}=\lambda_{A'_2B_2}=+1$ -- and (3) {\it The Tiger Whinnies}, (4) {\it The Cat Growls} -- and in case one of these is chosen we put $\lambda_{A'_1B_2}=\lambda_{A'_2B_1}=-1$. For the measurement $A'B'$ participants choose among (1) {\it The Tiger Snorts}, (2) {\it The Cat Meows} -- and in case one of these is chosen we put $\lambda_{A'_1B'_1}=\lambda_{A'_2B'_2}=+1$ -- and (3) {\it The Tiger Meows}, (4) {\it The Cat Snorts} -- and in case one of these is chosen we put $\lambda_{A'_1B'_2}=\lambda_{A'_2B'_1}=-1$.

We now evaluate the expectation values $E(A', B')$, $E(A', B)$, $E(A, B')$ and $E(A,B)$ associated with the coincidence experiments $A'B'$, $A'B$, $AB'$ and $AB$, respectively, and substitute these values into the Clauser-Horne-Shimony-Holt (CHSH) version of Bell's inequality \cite{chsh69}
\begin{equation} \label{chsh}
-2 \le E(A',B')+E(A',B)+E(A,B')-E(A,B) \le 2
\end{equation}
One typically says that, if Eq. (\ref{chsh}) is violated, a classical Kolmogorovian probabilistic description of the data is not possible \cite{aerts1986,af1982,pitowsky1989}. In analogy with the quantum violation of the CHSH inequality, we call the phenomenon `cognitive entanglement' between the given concepts, and remark that it is the necessary appearance of non-product structures in the explicit quantum-theoretic model in Sec. \ref{animalactsquantum} 
that in our approach justifies this naming. 

We actually performed a test involving 81 subjects who were presented with a form to be filled out in which they were asked to choose among the above alternatives in experiments $AB$, $A'B$, $AB'$ and $A'B'$. If we denote by $p(A_1,B_1)$, $p(A_1,B_2)$, $p(A_2,B_1)$, $p(A_2,B_2)$, the probability that {\it The Horse Growls}, {\it The Bear Whinnies},  
{\it The Horse Whinnies}, {\it The Bear Growls}, respectively, is chosen in the coincidence experiment $AB$, and so on in the other experiments, these probabilities are $p(A_1,B_1)=0.049$, $p(A_1,B_2)=0.630$, $p(A_2,B_1)=0.259$, $p(A_2,B_2)=0.062$, in experiment $AB$,
$p(A_1,B'_1)=0.593$, $p(A_1,B'_2)=0.025$, $p(A_2,B'_1)=0.296$, $p(A_2,B'_2)=0.086$, in experiment $AB'$,
$p(A'_1,B_1)=0.778$, $p(A'_1,B_2)=0.086$, $p(A'_2,B_1)=0.086$, $p(A'_2,B_2)=0.049$, in experiment $A'B$,
$p(A'_1,B'_1)=0.148$, $p(A'_1,B'_2)=0.086$, $p(A'_2,B'_1)=0.099$, $p(A'_2,B'_2)=0.667$, in experiment $A'B'$. Therefore, the expectation values are $E(A,B)=p(A_1,B_1)-p(A_1,B_2)-p(A_2,B_1)+p(A_2,B_2)=-0.7778$, $E(A,B')=p(A_1,B'_1)-p(A_1,B'_2)-p(A_2,B'_1)+p(A_2,B'_2)=0.3580$, $E(A',B)=p(A'_1,B_1)-p(A'_1,B_2)-p(A'_2,B_1)+p(A'_2,B_2)=0.6543$, $E(A',B')=p(A'_1,B'_1)-p(A'_1,B'_2)-p(A'_2,B'_1)+p(A'_2,B'_2)=0.6296$. 

Hence, Eq. (\ref{chsh}) gives $E(A',B')+E(A',B)+E(A,B')-E(A,B)=2.4197$, which is significantly greater than 2. This violation is close to Tsirelson's bound \cite{tsirelson80}, the maximal quantum violation of $2\sqrt{2}$, in case only product measurements are considered, so that it does reveal 
the presence of genuine entanglement 
in the situation considered with {\it The Animal Acts}, 
as we will see in the next section.

\subsection{A quantum representation in complex Hilbert space\label{animalactsquantum}}
Let us now construct a quantum representation in complex Hilbert space for the collected data by starting from an operational description of the conceptual entity {\it The Animal Acts}. The entity {\it The Animal Acts} is abstractly described by an initial state $p$. Measurement $AB$ has four outcomes $\lambda_{A_1B_1}$, $\lambda_{A_1B_2}$, $\lambda_{A_2B_1}$ and $\lambda_{A_2B_2}$, and four final states $p_{A_1B_1}$, $p_{A_1B_2}$, $p_{A_2B_1}$ and $p_{A_2B_2}$. Measurement $AB'$ has four outcomes $\lambda_{A_1B'_1}$, $\lambda_{A_1B'_2}$, $\lambda_{A_2B'_1}$ and $\lambda_{A_2B'_2}$, and four final states $p_{A_1B'_1}$, $p_{A_1B'_2}$, $p_{A_2B'_1}$ and $p_{A_2B'_2}$. Measurement $A'B$ has four outcomes $\lambda_{A'_1B_1}$, $\lambda_{A'_2B_1}$, $\lambda_{A'_1B_2}$ and $\lambda_{A'_2B_2}$, and four final states $p_{A'_1B_1}$, $p_{A'_1B_2}$, $p_{A'_2B_1}$ and $p_{A'_2B_2}$. Measurement $A'B'$ has four outcomes $\lambda_{A'_1B'_1}$, $\lambda_{A'_2B'_1}$, $\lambda_{A'_1B'_2}$ and $\lambda_{A'_2B'_2}$, and four final states $p_{A'_1B'_1}$, $p_{A'_1B'_2}$, $p_{A'_2B'_1}$ and $p_{A'_2B'_2}$.
Then, we consider the Hilbert space ${\mathbb C}^4$ as the state space of {\it The Animal Acts} and represent the state $p$ by the unit vector $|p\rangle \in {\mathbb C}^4$. We assume that $\{|p_{A_1B_1}\rangle, |p_{A_1B_2}\rangle, |p_{A_2B_1}\rangle,$ $ |p_{A_2B_2}\rangle \}$, $\{|p_{A_1B'_1}\rangle, |p_{A_1B'_2}\rangle, |p_{A_2B'_1}\rangle, |p_{A_2B'_2}\rangle\}$, $\{|p_{A'_1B_1}\rangle$, $ |p_{A'_1B_2}\rangle$, $|p_{A'_2B_1}\rangle$, $|p_{A'_2B_2}\rangle\}$, $\{|p_{A'_1B'_1}\rangle,$ $|p_{A'_1B'_2}\rangle, |p_{A'_2B'_1}\rangle, |p_{A'_2B'_2}\rangle\}$ are orthonormal (ON) bases of ${\mathbb C}^4$. Therefore, $|\langle p_{A_1B_1}|\psi\rangle|^2=p(A_1B_1)$, $|\langle p_{A_1B_2}|\psi\rangle|^2=p(A_1B_2)$, $|\langle p_{A_2B_1}|\psi\rangle|^2=p(A_2B_1)$, $|\langle p_{A_2B_2}|\psi\rangle|^2=p(A_2B_2)$, in the measurement $AB$. We proceed analogously for the other probabilities. Hence, the self-adjoint operators 
\begin{eqnarray}
{\cal E}_{AB}&=&\lambda_{A_1B_1}|p_{A_1B_1}\rangle \langle p_{A_1B_2}|+\lambda_{A_1B_2}|p_{A_1B_2}\rangle \langle p_{A_1B_1}| \nonumber \\
&&+\lambda_{A_2B_1}|p_{A_2B_1}\rangle \langle p_{A_2B_1}|+\lambda_{A_2B_2}|p_{A_2B_2}\rangle \langle p_{A_2B_2}| \nonumber\\
{\cal E}_{AB'}&=&\lambda_{A_1B'_1}|p_{A_1B'_1}\rangle \langle p_{A_1B'_2}|+\lambda_{A_1B'_2}|p_{A_1B'_2}\rangle \langle p_{A_1B'_1}| \nonumber \\
&&+\lambda_{A_2B'_1}|p_{A_2B'_1}\rangle \langle p_{A_2B'_1}|+\lambda_{A_2B'_2}|p_{A_2B'_2}\rangle \langle p_{A_2B'_2}| \nonumber\\
{\cal E}_{A'B}&=&\lambda_{A'_1B_1}|p_{A'_1B_1}\rangle \langle p_{A'_1B_2}|+\lambda_{A'_1B_2}|p_{A'_1B_2}\rangle \langle p_{A'_1B_1}| \nonumber \\
&&+\lambda_{A'_2B_1}|p_{A'_2B_1}\rangle \langle p_{A'_2B_1}|+\lambda_{A'_2B_2}|p_{A'_2B_2}\rangle \langle p_{A'_2B_2}| \nonumber\\
{\cal E}_{A'B'}&=&\lambda_{A'_1B'_1}|p_{A'_1B'_1}\rangle \langle p_{A'_1B'_2}|+\lambda_{A'_1B'_2}|p_{A'_1B'_2}\rangle \langle p_{A'_1B'_1}| \nonumber \\
&&+\lambda_{A'_2B'_1}|p_{A'_2B'_1}\rangle \langle p_{A'_2B'_1}|+\lambda_{A'_2B'_2}|p_{A'_2B'_2}\rangle \langle p_{A'_2B'_2}|
\end{eqnarray}
represent the measurements $AB$, $AB'$, $A'B$ and $A'B'$ in ${\mathbb C}^4$, respectively.

Let now the state $p$ of {\it The Animal Acts} be the entangled state represented by the unit vector $|p\rangle=|0.23e^{i13.93^\circ}, 0.62e^{i16.72^\circ},0.75e^{i9.69^\circ},0e^{i194.15^\circ}\rangle$ in the canonical basis of ${\mathbb C}^{4}$. This choice is not arbitrary, but deliberately `as close as possible to a situation of only product measurements', as we explain in \cite{asIQSA2012,asQI2013}. Moreover, we choose the outcomes $\lambda_{A_1B_1}$, \ldots, $\lambda_{A'_2B'_2}$ to be $\pm 1$, as in Sec. \ref{animalactsdescription}. In this case, we have proved in Ref. \cite{asIQSA2012} that 
\small
\begin{eqnarray}
{\cal E}_{AB}
&=&
\left( \begin{array}{cccc}
0.952 & -0.207-0.030i	&	0.224+0.007i & 0.003-0.006i \\				
-0.207+0.030i & -0.930	&	0.028-0.001i	&	-0.163+0.251i \\				
0.224-0.007i & 0.028+0.001i & -0.916 & -0.193+0.266i \\
0.003+0.006i & -0.163-0.251i & -0.193-0.266i & 0.895				
\end{array} \right)
\\
{\cal E}_{AB'}
&=&
\left( \begin{array}{cccc}
-0.001	&				0.587+0.397i &	0.555+0.434i &	0.035+0.0259i	\\			
0.587-0.397i & -0.489 &	0.497+0.0341i &	-0.106-0.005i \\	
0.555-0.434i & 0.497-0.0341i & -0.503	&	0.045-0.001i \\				
0.035-0.0259i &	-0.106+0.005i &	0.045+0.001i & 0.992	\end{array} \right)
\\
{\cal E}_{A'B}
&=&
\left( \begin{array}{cccc}
-0.587 &	0.568+0.353i	&	0.274+0.365i	&	0.002+0.004i \\																							
0.568-0.353i & 0.090	 &				0.681+0.263i & -0,110-0.007i \\				
0.274-0.365i &		0.681-0.263i & -0.484	&	0.150-0.050i \\				
0,002-0.004i & -0,110+0.007i & 0.150+0.050i &	0.981				
\end{array} \right)
\end{eqnarray}
\begin{eqnarray}
{\cal E}_{A'B'} 
&=&
\left( \begin{array}{cccc}
0.854	&				0.385+0.243i & -0.035-0.164i &	-0.115-0.146i \\				
0.385-0.243i & -0.700	&	0.483+0.132i & -0.086+0.212i \\				
-0.035+0.164i &	0.483-0.132i & 0.542 &	0.093+0.647i \\				
-0.115+0.146i &	-0.086-0.212i &	0.093-0.647i &	-0.697	
\end{array} \right)
\end{eqnarray}
\normalsize
Our quantum-theoretic modeling in the Hilbert space ${\mathbb C}^{4}$ of our cognitive test is completed. By recalling the following canonical isomorphisms, ${\mathbb C}^{4}\cong{\mathbb C}^{2}\otimes{\mathbb C}^{2}$ and $L({\mathbb C}^{4})\cong L({\mathbb C}^{2})\otimes L({\mathbb C}^{2})$, and the definitions of entangled states and measurements in Refs. \cite{asIQSA2012,asQI2013}, it can be proved that all measurements $AB$, $AB'$, $A'B$ and $A'B'$ are entangled with this choice of the entangled state. Moreover, the marginal distribution law is violated by all measurements, e.g., $p(A_1B_1)+p(A_1B_2) \ne p(A_1B'_1)+p(A_1B'_2)$. Since we are under Tsirelson's bound, this modeling is an example of a `nonlocal non-marginal box modeling 1', following the classification we have proposed in Ref. \cite{asQI2013}.

\section{The vessels of water entity \label{vesselsdescription}}
We have seen in Sec. \ref{animalacts} that there are unexpected connections between how a sentient human being connects conceptual entities through meaning in cognitive tests and how microscopic quantum entities are connected in entangled states in space-like separated spin experiments. Both kinds of entities violate Bell's inequalities and present entanglement. In this section, we consider a macroscopic entity, namely, `two vessels of water connected by a tube', which behaves in an analogous way. We believe that the `connected vessels of water example' still is a very good example because it provides an intuitive insight into `what entanglement is about', i.e. what conditions are necessary and sufficient for entanglement to manifest itself in reality, irrespective of whether it is physical or cognitive reality.
We came upon this example many years ago, when we were demonstrating how Bell's inequalities can be violated by ordinary macroscopic material entities by different examples \cite{aerts1982,aerts1985a,aerts1985b,aerts1991,aertsaertsbroekaertgabora2000}, and we will discuss it in some detail here.

We consider two vessels $V_A$ and $V_B$ connected by a tube $T$, containing a total of 20 liters of transparent water. Coincidence experiments $A$ and $B$ consist in siphons $S_A$ and $S_B$ pouring out water from vessels $V_A$ and $V_B$, respectively, and collecting the water in reference vessels $R_A$ and $R_B$, where the volume of collected water is measured. If more than 10 liters are collected for experiments $A$ or $B$ we put $E(A)=+1$ or $E(B)=+1$, respectively, and if fewer than 10 liters are collected for experiments $A$ or $B$, we put $E(A)=-1$ or $E(B)=-1$, respectively. We define experiments $A'$ and $B'$, which consist in taking a small spoonful of water out of the left vessel and the right vessel, respectively, and verifying whether the water is transparent. We have $E(A')=+1$ or $E(A')=-1$, depending on whether the water in the left vessel turns out to be transparent or not, and $E(B')=+1$ or $E(B')=-1$, depending on whether the water in the right vessel turns out to be transparent or not. We put $E(AB)=+1$ if $E(A)=+1$ and $E(B)=+1$ or $E(A)=-1$ and $E(B)=-1$, and $E(AB)=-1$ if $E(A)=+1$ and $E(B)=-1$ or $E(A)=-1$ and $E(B)=+1$, if the coincidence experiment $AB$ is performed. We can thus define the expectation value $E(A,B)$ for the coincidence experiment $AB$ in a traditional way. Similarly, we put $E(A'B)=+1$ if $E(A')=+1$ and $E(B)=+1$ or $E(A')=-1$ and $E(B)=-1$ and the coincidence experiment $A'B$ is performed. And we have $E(AB')=+1$ if $E(A)=+1$ and $E(B')=+1$ or $E(A)=-1$ and $E(B')=-1$ and the coincidence experiment $AB'$ is performed, and further $E(A'B')=+1$ if $E(A')=+1$ and $E(B')=+1$ or $E(A')=-1$ and $E(B')=-1$ and the coincidence experiment $A'B'$ is performed. Hence, we can define the expectation values  $E(A',B)$, $E(A,B')$ and $E(A',B')$ corresponding to the coincidence experiments $A'B$, $AB'$ and $A'B'$, respectively. Now, since each vessel contains 10 liters of transparent water, we find that these expectation values are $E(A, B)=-1$, $E(A', B)=+1$, $E(A, B')=+1$ and $E(A', B')=+1$, which gives $E(A',B')+E(A',B)+E(A,B')-E(A,B)=+4$. This is the maximum possible violation of the CHSH form of Bell's inequalities.

There are deep structural and conceptual connections between the cognitive and physical situations violating Bell's inequalities. The main reason why these interconnected water vessels can violate Bell's inequalities is because the water in the vessels has not yet been subdivided into two volumes before the measurement starts. The water in the vessels is only `potentially' subdivided into volumes whose sum is 20 liters. It is not until the measurement is actually carried out that one of these potential subdivisions actualizes, i.e. one part of the 20 liters is collected in reference vessel $R_A$ and the other part is collected in reference vessel $R_B$. This is very similar to the combination of concepts {\it The Animal Acts} in Sec. \ref{animalacts} not collapsing into one of the four possibilities {\it The Horse Growls}, {\it The Bear Whinnies}, {\it The Bear Growls} or {\it The Bear Whinnies} before the coincidence measurement $AB$ starts. It is the coincidence measurement itself which makes the combination {\it The Animal Acts} collapse into one of these four possibilities. The same holds for the interconnected water vessels. The coincidence experiment $AB$ with the siphons is what causes the total volume of 20 liters of water to be split into two volumes, and it is this which creates the correlation for $AB$ giving rise to $E(A,B)=-1$. It can easily be calculated that if we take away the tube and suppose that, before the measurement, the water is already subdivided over the two vessels, which are now no longer interconnected, although still an anti-correlation would be measured for the coincidence experiments between $A$ and $B$, the perfect correlations between $A$ and $B'$, and between $A'$ and $B$ no longer hold, one of them changing into an anti-correlation. This makes that Bell's inequality is no longer violated, i.e. $E(A',B')+E(A',B)+E(A,B')-E(A,B)=+2$ -- by the way, this is also true in general when the initial state of the vessels of water is a mixture of product states. This proves that the tube, provoking the `potentiality of the anti-correlation for $A$ and $B$', is essential for Bell's inequality to be violated. In {\it The Animal Acts}, it is the presence of `meaning' in the mind of the choosing person that is necessary to provoke a violation of Bell's inequalities. In the vessels of water, it is the presence of `water' in the two connected vessels which is necessary to provoke a violation of Bell's inequalities.

\section{A quantum representation of the vessels of water\label{vessels}}
In this section, we elaborate a Hilbert space representation for the vessels of water situation: this result is 
new and was not investigated neither expected when this example was originally conceived.
 
Let us provide a preliminary description of the experiments with the vessels of water, as in Sec. \ref{animalactsquantum}. The entity `vessels of water' is abstractly described each time by a state $p$. Measurement $AB$ has four outcomes, $\lambda_{A_{1}B_{1}}$, $\lambda_{A_{1}B_{2}}$, $\lambda_{A_{2}B_{1}}$ and $\lambda_{A_{2}B_{2}}$, and four final states, $p_{A_{1}B_{1}}$, $p_{A_{1}B_{2}}$, $p_{A_{2}B_{1}}$ and $p_{A_{2}B_{2}}$. Measurement $AB'$ has four outcomes, $\lambda_{A_{1}B'_{1}}$, $\lambda_{A_{1}B'_{2}}$, $\lambda_{A_{2}B'_{1}}$ and $\lambda_{A_{2}B'_{2}}$, and four final states, $p_{A_{1}B'_{1}}$, $p_{A_{1}B'_{2}}$, $p_{A_{2}B'_{1}}$ and $p_{A_{2}B'_{2}}$.
Measurement $A'B$ has four outcomes $\lambda_{A'_{1}B_{1}}$, $\lambda_{A'_{1}B_{2}}$, $\lambda_{A'_{2}B_{1}}$ and $\lambda_{A'_{2}B_{2}}$, and four final states $p_{A'_{1}B_{1}}$, $p_{A'_{1}B_{2}}$, $p_{A'_{2}B_{1}}$ and $p_{A'_{2}B_{2}}$.
Measurement $A'B'$ has four outcomes $\lambda_{A'_{1}B'_{1}}$, $\lambda_{A'_{1}B'_{2}}$, $\lambda_{A'_{2}B'_{1}}$ and $\lambda_{A'_{2}B'_{2}}$, and four final states $p_{A'_{1}B'_{1}}$, $p_{A'_{1}B'_{2}}$, $p_{A'_{2}B'_{1}}$ and $p_{A'_{2}B'_{2}}$.

To work out a quantum-mechanical model in the Hilbert space ${\mathbb C}^4$ for the vessels of water situation, we consider the entangled state $p$ represented by the unit vector $|p\rangle=|0, \sqrt{0.5}e^{i\alpha}, \sqrt{0.5}e^{i\beta}, 0\rangle$ as describing the vessels of water situation with transparent water, and represent the measurement $AB$ by the ON (canonical) basis
$|p_{A_{1}B_{1}}\rangle=|1, 0, 0, 0\rangle$, $|p_{A_{1}B_{2}}\rangle=|0, 1, 0, 0\rangle$
$|p_{A_{2}B_{1}}\rangle=|0, 0, 1, 0\rangle$, $|p_{A_{2}B_{2}}\rangle=|0, 0, 0, 1\rangle$.
This gives indeed the correct probabilities in the state $p$, that is,
$p(\lambda_{A_{1}B_{1}})=|\langle p_{A_{1}B_{1}}|p\rangle|^2=0$,
$p(\lambda_{A_{1}B_{2}})=|\langle p_{A_{1}B_{2}}|p\rangle|^2=0.5$,
$p(\lambda_{A_{2}B_{1}})=|\langle p_{A_{2}B_{1}}|p\rangle|^2=0.5$,
$p(\lambda_{A_{2}B_{2}})=|\langle p_{A_{2}B_{2}}|p\rangle|^2=0$.
In the coincidence measurement $AB'$, we take the ON basis
$|p_{A_{1}B'_{1}}\rangle=$ $|0, \sqrt{0.5}e^{i\alpha}, \sqrt{0.5}e^{i\beta}, 0\rangle$, 
$|p_{A_{1}B'_{2}}\rangle=|0, \sqrt{0.5}e^{i\alpha}, -\sqrt{0.5}e^{i\beta}, 0\rangle$,
$|p_{A_{2}B'_{1}}\rangle=| 1, 0, 0, 0\rangle$,
$|p_{A_{2}B'_{2}}\rangle=|0, 0, 0, 1\rangle$.  
We have the correct probabilities in the state $p$, that is,
$p(\lambda_{A_{1}B'_{1}})=|\langle p_{A_{1}B'_{1}}|p\rangle|^2=1$,
$p(\lambda_{A_{1}B'_{2}})=|\langle p_{A_{1}B'_{2}}|p\rangle|^2=0$,
$p(\lambda_{A_{2}B'_{1}})=|\langle p_{A_{2}B'_{1}}|p\rangle|^2=0$, 
$p(\lambda_{A_{2}B'_{2}})=|\langle p_{A_{2}B'_{2}}|p\rangle|^2=0$.
In the coincidence measurement $A'B$, we choose the ON basis 
$|p_{A'_{1}B_{1}}\rangle=|0, \sqrt{0.5}e^{i\alpha}, \sqrt{0.5}e^{i\beta}, 0\rangle$,
$|p_{A'_{1}B_{2}}\rangle= | 1, 0, 0, 0\rangle$, 
$|p_{A'_{2}B_{1}}\rangle= |0, \sqrt{0.5}e^{i\alpha},$ $-\sqrt{0.5}e^{i\beta}, 0\rangle$, 
$|p_{A'_{2}B_{2}}\rangle= |0, 0, 0, 1\rangle$.  
As expected, we get probability 1 for the outcome $\lambda_{A'_{1}B_{1}}$ in the state $p$.
In the coincidence measurement $A'B'$, we take the ON basis
$|p_{A'_{1}B'_{1}}\rangle=|0, \sqrt{0.5}e^{i\alpha}, \sqrt{0.5}e^{i\beta}, 0\rangle$,
$|p_{A'_{1}B'_{2}}\rangle= | 1, 0, 0, 0\rangle$, 
$|p_{A'_{2}B'_{1}}\rangle= |0, 0, 0, 1\rangle$,
$|p_{A'_{2}B'_{2}}\rangle= |0, \sqrt{0.5}e^{i\alpha}, -\sqrt{0.5}e^{i\beta}, 0\rangle$.  
As expected, we get probability 1 for the outcome $\lambda_{A'_{1}B'_{1}}$ in the state $p$.

Let us now explicitly construct the self-adjoint operators representing the measurements $AB$, $AB'$, $A'B$ and $A'B'$. They are respectively given by
\small 
\begin{eqnarray}
&{\mathcal E}_{AB}=\sum_{i,j=1}^{2}\lambda_{A_{i}B_{j}}|p_{A_{i}B_{j}}\rangle \langle p_{A_{i}B_{j}}|
=
\left( \begin{array}{cccc}
\lambda_{A_{1}B_{1}} & 0 & 0 & 0 \\
0 & \lambda_{A_{1}B_{2}} & 0 & 0 \\
0 & 0 & \lambda_{A_{2}B_{1}} & 0 \\
0 & 0 & 0 & \lambda_{A_{2}B_{2}}
\end{array} \right) \label{1} \\
&{\mathcal E}_{AB'}=\sum_{i,j=1}^{2}\lambda_{A_{i}B'_{j}}|p_{A_{i}B'_{j}}\rangle \langle p_{A_{i}B'_{j}}| \nonumber \\
&=
\left( \begin{array}{cccc}
\lambda_{A_{2}B'_{1}} & 0 & 0 & 0 \\
0 & 0.5(\lambda_{A_{1}B'_{1}}+\lambda_{A_{1}B'_{2}}) & 0.5e^{i(\alpha-\beta)}(\lambda_{A_{1}B'_{1}}-\lambda_{A_{1}B'_{2}}) & 0 \\
0 & 0.5e^{-i(\alpha-\beta)}(\lambda_{A_{1}B'_{1}}-\lambda_{A_{1}B'_{2}}) & 0.5(\lambda_{A_{1}B'_{1}}+\lambda_{A_{1}B'_{2}}) & 0 \\
0 & 0 & 0 & \lambda_{A_{2}B'_{2}}
\end{array} \right) \label{2} \\
&{\mathcal E}_{A'B}=\sum_{i,j=1}^{2}\lambda_{A'_{i}B_{j}}|p_{A'_{i}B_{j}}\rangle \langle p_{A'_{i}B_{j}}| \nonumber
\end{eqnarray}
\begin{eqnarray}
&=
\left( \begin{array}{cccc}
\lambda_{A'_{1}B_{2}} & 0 & 0 & 0 \\
0 & 0.5(\lambda_{A'_{1}B_{1}}+\lambda_{A'_{2}B_{1}}) & 0.5e^{i(\alpha-\beta)}(\lambda_{A'_{1}B_{1}}-\lambda_{A'_{2}B_{1}}) & 0 \\
0 & 0.5e^{-i(\alpha-\beta)}(\lambda_{A'_{1}B_{1}}-\lambda_{A'_{2}B_{1}}) & 0.5(\lambda_{A'_{1}B_{1}}+\lambda_{A'_{2}B_{1}}) & 0 \\
0 & 0 & 0 & \lambda_{A'_{2}B_{2}}
\end{array} \right) \label{3}
\\
&{\mathcal E}_{A'B'}=\sum_{i,j=1}^{2}\lambda_{A'_{i}B'_{j}}|p_{A'_{i}B'_{j}}\rangle \langle p_{A'_{i}B'_{j}}| \nonumber \\
&=
\left( \begin{array}{cccc}
\lambda_{A'_{1}B'_{2}} & 0 & 0 & 0 \\
0 & 0.5(\lambda_{A'_{1}B'_{1}}+\lambda_{A'_{2}B'_{2}}) & 0.5e^{i(\alpha-\beta)}(\lambda_{A'_{1}B'_{1}}-\lambda_{A'_{2}B'_{2}}) & 0 \\
0 & 0.5e^{-i(\alpha-\beta)}(\lambda_{A'_{1}B'_{1}}-\lambda_{A'_{2}B'_{2}}) & 0.5(\lambda_{A'_{1}B'_{1}}+\lambda_{A'_{2}B'_{2}}) & 0 \\
0 & 0 & 0 & \lambda_{A'_{2}B'_{1}}
\end{array} \right) \label{4}
\end{eqnarray}
\normalsize
The self-adjoint operators corresponding to measuring the expectation values are instead obtained by putting 
$\lambda_{A_{i}B_{i}}=\lambda_{A_{i}B'_{i}}=\lambda_{A'_{i}B_{i}}=\lambda_{A'_{i}B'_{i}}=+1$, $i=1,2$ and $\lambda_{A_{i}B_{j}}=\lambda_{A_{i}B'_{j}}=\lambda_{A'_{i}B_{j}}=\lambda_{A'_{i}B'_{j}}=-1$, $i,j=1,2;i \ne j$, as in our experiment in Sec. \ref{vesselsdescription}. If we now insert these values into Eqs. (\ref{1})--(\ref{4}) and define the `Bell operator' as
\small
\begin{equation}
B={\mathcal E}_{A'B'}+{\mathcal E}_{A'B}+{\mathcal E}_{AB'}-{\mathcal E}_{AB}=
\left( \begin{array}{cccc}
0 & 0 & 0 & 0 \\
0 & 2 & 2e^{i(\alpha-\beta)} & 0 \\
0 & 2e^{-i(\alpha-\beta)} & 2 & 0 \\
0 & 0 & 0 & 0
\end{array} \right)
\end{equation}
\normalsize
its expectation value in the entangled state $p$ is
\small
\begin{eqnarray}
&\langle p|B|p\rangle \nonumber \\
&=
\left( \begin{array}{cccc}
0 & \sqrt{0.5}e^{-i\alpha} & \sqrt{0.5}e^{-i\beta} & 0 \\
\end{array} \right)\left( \begin{array}{cccc}
0 & 0 & 0 & 0 \\
0 & 2 & 2e^{i(\alpha-\beta)} & 0 \\
0 & 2e^{-i(\alpha-\beta)} & 2 & 0 \\
0 & 0 & 0 & 0
\end{array} \right)\left( \begin{array}{c}
0   \\
\sqrt{0.5}e^{i\alpha} \\
\sqrt{0.5}e^{i\beta} \\
0 
\end{array} \right)=4 
\end{eqnarray}
\normalsize
which gives the same value in the CHSH inequality as in Sec. \ref{vesselsdescription}. A completely analogous construction can be performed if the entangled state $q$ represented by the unit vector $|q\rangle=|0, \sqrt{0.5}e^{i\alpha}, -\sqrt{0.5}e^{i\beta}, 0\rangle$ is chosen to describe the vessels with non-transparent water.

We add some conclusive remarks that are discussed in detail in Ref. \cite{asQI2013}. The measurement $AB$ is a product measurement, since it has the product states represented by the vectors in the canonical basis of ${\mathbb C}^{4}$ as final states. Indeed, $AB$ `divides' the water into two separated volumes of water, thus `destroying' entanglement, to arrive at a situation of a product state. The measurements $AB'$ and $A'B$ are instead entangled measurements, since they have the entangled states represented by $|0, \sqrt{0.5}e^{i\alpha}, \sqrt{0.5}e^{i\beta}, 0\rangle$ and $|0, \sqrt{0.5}e^{i\alpha}, -\sqrt{0.5}e^{i\beta}, 0\rangle$ as final states. Indeed, since all the water is poured out of the two vessels, the water has not been divided, and inside the reference vessel it keeps being a whole, i.e. entangled. If another two siphons are put in the reference vessel where all the water has been collected, the same experiment can be performed, violating Bell's inequalities. The measurement $A'B'$ has also entangled states as possible final states, hence it is entangled. This measurement leaves the vessels of water unchanged, hence it is naturally an entangled measurement. We finally observe that the marginal ditribution law is violated in the case of the vessels of water. Indeed, we have, e.g., $0.5=p(\lambda_{A_{1}B_{1}})+p(\lambda_{A_{1}B_{2}}) \ne p(\lambda_{{A_1}B'_{1}})+p(\lambda_{{A_1}B'_{2}})=1$. Since this vessels of water model violates Bell's inequalities beyond Tsirelson's bound, we can say that our ${\mathbb C}^{4}$ representation is an example of a `nonlocal non-marginal box modeling 2', if we follow the classification in Ref. \cite{asQI2013}.

\section{An alternative model for the vessels of water\label{alternativevessels}}
In this section, we provide an alternative model in ${\mathbb C}^{4}$ for the vessels of water situation. This model is interesting, in our opinion, because it serves to show that 
where entanglement is located, in the state, or on the level of the measurement, depends on the way in which the tensor product isomorphism with the compound entity Hilbert space is chosen. 

In the representation in Sec. \ref{vessels}, we have given preference to the first coincidence measurement $AB$ which we have chosen as a product, i.e. we have represented it by the canonical basis of ${\mathbb C}^4$. This means that the entanglement of this `state-measurement' situation has been completely put into the state. Let us identify this entanglement on the level of the probabilities by using Th. 2 in Ref. \cite{asQI2013}. We have $p(\lambda_{A_{1}B_{1}})=p(\lambda_{A_{2}B_{2}})=0$, $p(\lambda_{A_{1}B_{1}})=p(\lambda_{A_{2}B_{2}})=0.5$ in both states $p$ and $q$.
Suppose that we search numbers $a,b,a',b'\in [0,1]$ such that $p(\lambda_{A_{1}B_{1}})=a\cdot b$, $p(\lambda_{A_{1}B_{2}})=a \cdot b'$, $p(\lambda_{A_{2}B_{1}})=a' \cdot b$ and $p(\lambda_{A_{2}B_{2}})=a' \cdot b'$. Then, we get that $a=0$ or $b=0$. Since $a \cdot b'=0.5$, we cannot have that $a=0$, and hence $b=0$. But then $a' \cdot b$ cannot be equal to $0.5$. This entails that the probabilities do not compose into a product, hence there is entanglement in the considered `state-measurement' situation, this entanglement being `a joint property of state and measurement', and not of one apart.

Let us now consider the probabilities of the measurement $AB'$. We have $p(\lambda_{A_{1}B'_{1}})=1$, $p(\lambda_{A_{1}B'_{2}})=p(\lambda_{A_{2}B'_{1}})=p(\lambda_{A_{2}B'_{2}})=0$ in the state $p$. We can again look for numbers $a,b,a',b' \in [0,1]$ such that
$p(\lambda_{A_{1}B'_{1}})=a\cdot b$, $p(\lambda_{A_{1}B'_{2}})=a \cdot b'$, $p(\lambda_{A_{2}B'_{1}})=a' \cdot b$ and $\quad p(\lambda_{A_{2}B'_{1}})=a' \cdot b'$.
We find the solution $a'=b'=0$, and $a=b=1$, which is unique. Indeed, from $a \cdot b =1$ follows that $a\not=0$ and $b\not=0$, and hence from $a \cdot b'=0$ and $a' \cdot b=0$ follows then $a'=b'=0$. This implies that we could model this `state-measurement' situation by a product state and a product measurement. Let us do this explicitly in ${\mathbb C}^4$. If this time we represent the state $p'$ with transparent water by the unit vector $|p'\rangle= |1,0,0,0\rangle$, and the measurement $AB'$ in the canonical basis, we get the wanted result. This also implies that we have single probabilities $p(\lambda_{A_{1}})$, $p(\lambda_{A_{2}})$, $p(\lambda_{B'_{1}})$ and $p(\lambda_{B'_{2}})$ such that $p(\lambda_{A_{1}})=p(\lambda_{A_{2}})=1$, $p(\lambda_{B'_{1}})=p(\lambda_{B'_{2}})=0$. We can construct also the second and third `state measurement' in the same space, and with the same state. It gives $p(\lambda_{A'_{1}})=p(\lambda_{B_{1}})=1$, $p\lambda_{A'_{2}})=p(\lambda_{B_{2}})=0$. 

Proceeding in this way, we can propose an alternative quantum model where we use the product state $p'$, represented by $|1,0,0,0 \rangle$, and the product measurements $AB'$, $A'B$ and $A'B'$, all represented by the canonical ON basis in ${\mathbb C}^{4}$. Only $AB$ is entangled in this construction and corresponds to the ON set 
$|p'_{1}\rangle=|0,1,0,0\rangle$, $|p'_{2}\rangle=|\sqrt{0.5}e^{i\alpha}, 0, 0, \sqrt{0.5}e^{i\beta}\rangle$, 
$|p'_{3}\rangle=|\sqrt{0.5}e^{i\alpha}, 0, 0, -\sqrt{0.5}e^{i\beta}\rangle$, 
$|p'_{4}\rangle=|0,0,1,0\rangle$, 
as one can verify at once. This gives rise to the self-adjoint operators
\small
\begin{eqnarray}
{\mathcal E'}_{AB}=\lambda_{A_{1}B_{1}}|p_{1}\rangle\langle p_{1}|+\lambda_{A_{1}B_{2}}|p_{2}\rangle\langle p_{2}|+\lambda_{A_{2}B_{1}}|p_{3}\rangle\langle p_{3}|+\lambda_{A_{2}B_{2}}|p_{4}\rangle\langle p_{4}| \nonumber \\
=
\left( \begin{array}{cccc}
0.5(\lambda_{A_{1}B_{1}}+\lambda_{A_{2}B_{1}}) & 0 & 0 & 0.5e^{i(\alpha-\beta)}(\lambda_{A_{1}B_{2}}-\lambda_{A_{2}B_{1}}) \\
0 & \lambda_{A_{1}B_{1}} & 0 & 0 \\
0 & 0 & \lambda_{A_{2}B_{2}} & 0 \\
0.5e^{-i(\alpha-\beta)}(\lambda_{A_{1}B_{2}}-\lambda_{A_{2}B_{1}}) & 0 & 0 & 0.5(\lambda_{A_{1}B_{2}}+\lambda_{A_{2}B_{1}})
\end{array} \right) \label{1'}
\end{eqnarray}
\begin{eqnarray}
&{\mathcal E'}_{AB'}=\left( \begin{array}{cccc}
\lambda_{A_{1}B'_{1}} & 0 & 0 & 0 \\
0 & \lambda_{A_{1}B'_{2}} & 0 & 0 \\
0 & 0 & \lambda_{A_{2}B'_{1}} & 0 \\
0 & 0 & 0 & \lambda_{A_{2}B'_{2}}
\end{array} \right)
\quad
{\mathcal E'}_{A'B}=\left( \begin{array}{cccc}
\lambda_{A'_{1}B_{1}} & 0 & 0 & 0 \\
0 & \lambda_{A'_{1}B_{2}} & 0 & 0 \\
0 & 0 & \lambda_{A'_{2}B_{1}} & 0 \\
0 & 0 & 0 & \lambda_{A'_{2}B_{2}}
\end{array} \right) \label{2'}
\\
& {\mathcal E'}_{A'B'}=\left( \begin{array}{cccc}
\lambda_{A'_{1}B'_{1}} & 0 & 0 & 0 \\
0 & \lambda_{A'_{1}B'_{2}} & 0 & 0 \\
0 & 0 & \lambda_{A'_{2}B'_{1}} & 0 \\
0 & 0 & 0 & \lambda_{A'_{2}B'_{2}}
\end{array} \right) \label{3'}
\end{eqnarray}
\normalsize
As usual, if we measure expectation values, i.e. the outcomes are all either $+1$ or $-1$, and insert them into Eqs. (\ref{1'})--(\ref{3'}), we can directly calculate the expectation values in the state $p'$. We find
$\langle p' |{\mathcal E'}_{A'B'}+{\mathcal E'}_{A'B}+{\mathcal E'}_{AB'}-{\mathcal E'}_{AB}|p'\rangle=4$,
as expected. An analogous construction can be worked out for the state of the vessels with non-transparent water.


\begin{thebibliography}{99}
\bibitem{asIQSA2012} Aerts, D., Sozzo, S: Quantum Entanglement in Concept Combinations. Int. J. Theor. Phys. \emph{ArXiv:1302.3831 [cs.AI]} (2013)


\bibitem{bell1964} Bell, J.S.: On the Einstein-Podolsky-Rosen paradox. Physics. \textbf{1}, 195--200 (1964)

\bibitem{chsh69} Clauser, J.F., Horne, M.A., Shimony, A., Holt, R.A.: Proposed Experiment to Test Local Hidden-variable Theories. Phys. Rev. Lett. 23, 880--884 (1969)

\bibitem{aerts1986} Aerts, D.: A Possible Explanation for the Probabilities of Quantum Mechanics. J. Math. Phys. 27, 202--210 (1986)

\bibitem{af1982} Accardi, L., Fedullo, A.: On the Statistical Meaning of Complex Numbers in Quantum Theory. Lett. Nuovo Cim. 34, 161--172 (1982)

\bibitem{pitowsky1989} Pitowsky, I.: Quantum Probability, Quantum Logic. Lecture Notes in Physics vol. {\bf 321}.  Springer, Berlin (1989)

\bibitem{aerts1982} Aerts, D.: Example of a Macroscopical Situation That Violates Bell Inequalities. Lett. Nuovo Cim. 34, 107--111 (1982)

\bibitem{aerts1985a} Aerts, D.: The Physical Origin of the EPR Paradox and How to Violate Bell Inequalities by Macroscopical Systems. In: Lathi, P., Mittelstaedt, P. (eds.) Symposium on the Foundations of Modern Physics: 50 Years of the Einstein-Podolsky-Rosen Gedankenexperiment, pp. 305--320. World Scientific, Singapore (1985)

\bibitem{aerts1985b} Aerts, D.: A Possible Explanation for the Probabilities of Quantum Mechanics and a Macroscopical Situation That Violates Bell Inequalities. In: Mittelstaedt, P., Stachow, E.W. (eds.) Recent Developments in Quantum Logic, pp. 235--251. Bibliographisches Institut, Mannheim (1985) 

\bibitem{aerts1991} Aerts, D.: A Mechanistic Classical Laboratory Situation Violating the Bell Inequalities with 2$\sqrt{2}$, Exactly `in the Same Way' as its Violations by the EPR Experiments. Helv. Phys. Acta 64, 1--23 (1991)

\bibitem{aertsaertsbroekaertgabora2000} Aerts, D., Aerts, S., Broekaert, J., Gabora, L.: The Violation of Bell Inequalities in the Macroworld. Found. Phys. 30, 1387--1414 (2000)


\bibitem{aertsaerts95} Aerts, D., Aerts, S.: Applications of Quantum Statistics in Psychological Studies of Decision Processes. Found. Sci. 1, 85--97 (1995)

\bibitem{aertsgabora2005a} Aerts, D. Gabora, L.: A Theory of Concepts and Their Combinations I. The structure of the sets of contexts and properties. Kybernetes 34, 167--191 (2005)

\bibitem{aertsgabora2005b} Aerts, D. Gabora, L.: A Theory of Concepts and Their Combinations II. A Hilbert space representation. Kybernetes 34, 192--221 (2005)

\bibitem{bruzaetal2007} Bruza, P.D., Lawless, W., van Rijsbergen, C.J., Sofge, D., Editors: Proceedings of the AAAI Spring Symposium on Quantum Interaction, March 27--29. Stanford University, Stanford (2007)

\bibitem{bruzaetal2008} Bruza, P.D., Lawless, W., van Rijsbergen, C.J., Sofge, D., Editors: Quantum Interaction: Proceedings of the Second Quantum Interaction Symposium. College Publications, London (2008)

\bibitem{aerts2009} Aerts, D.: Quantum Structure in Cognition. J. Math. Psychol. 53, 314--348 (2009)

\bibitem{bruzaetal2009} Bruza, P.D., Sofge, D., Lawless, W., Van Rijsbergen, K., Klusch, M., Editors: Proceedings of the Third Quantum Interaction Symposium. Lecture Notes in Artificial Intelligence vol. \textbf{5494}. Springer, Berlin (2009)

\bibitem{pb2009} Pothos, E.M., Busemeyer, J.R.: A Quantum Probability Model Explanation for Violations of `Rational' Decision Theory. Proc. Roy. Soc. B 276, 2171--2178 (2009)

\bibitem{k2010} Khrennikov, A.Y.: Ubiquitous Quantum Structure. Springer, Berlin (2010)

\bibitem{songetal2011} Song, D., Melucci, M., Frommholz, I., Zhang, P., Wang, L., Arafat, S., Editors: Quantum Interaction. LNCS vol. {\bf 7052}. Springer, Berlin (2011)

\bibitem{bpft2011} Busemeyer, J.R., Pothos, E., Franco, R., Trueblood, J.S.: A Quantum Theoretical Explanation for Probability Judgment `Errors'. Psychol. Rev. 118, 193--218 (2011)

\bibitem{bb2012} Busemeyer, J.R., Bruza, P.D.: Quantum Models of Cognition and Decision. Cambridge University Press, Cambridge (2012)

\bibitem{busemeyeretal2012} Busemeyer, J. R., Dubois, F., Lambert-Mogiliansky, A., Melucci, M., Editors (2012). Quantum Interaction. LNCS vol. {\bf 7620}. Springer, Berlin (2012)

\bibitem{ags2012} Aerts, D., Gabora, L., S. Sozzo, S.: Concepts and Their Dynamics: A Quantum--theoretic Modeling of Human Thought. Top. Cogn. Sci. (in print). \emph{ArXiv: 1206.1069 [cs.AI]}

\bibitem{abgs2012} Aerts, D., Broekaert, J., Gabora, L., Sozzo, S.: Quantum Structure and Human Thought. Behav. Bra. Sci. 36, 274--276 (2013)

\bibitem{as2011} Aerts, D., Sozzo, S.: Quantum Structure in Cognition: Why and How Concepts are Entangled. LNCS vol. \textbf{7052}, 118--129. Springer, Berlin (2011)

\bibitem{asQI2013} Aerts, D, Sozzo, S.: The Entanglement Zoo I. Foundational and Structural Aspects, accepted in LNCS (2013)

\bibitem{aerts1999b} Aerts, D.: Foundations of Quantum Physics: A General Realistic and Operational Approach. Int. J. Theor. Phys. 38, 289--358 (1999)

\bibitem{aerts2009b} Aerts, D.: Quantum Particles as Conceptual Entities: A Possible Explanatory Framework for Quantum Theory. Found. Sci. 14, 361--411 (2010)

\bibitem{tsirelson80} Tsirelson, B.S.: Quantum Generalizations of Bell's Inequality. Lett. Math. Phys. 4, 93--100 (1980)

\end{thebibliography}
\end{document}